\documentclass[10pt, a4paper, conference, compsocconf]{IEEEtran}

\usepackage{times}
\usepackage{balance}
\usepackage{setspace}
\usepackage{algorithm}
\usepackage{algorithmic}
\usepackage{boxedminipage} 
\usepackage{graphicx}
\usepackage{verbatim} 
\usepackage{amsmath}
\usepackage{times} 
\usepackage{gensymb}
\usepackage{dsfont}
\usepackage{multirow}
\usepackage{amsfonts}

\usepackage[T1]{fontenc}
\usepackage{url}
\usepackage{amssymb}
\usepackage[tight,footnotesize]{subfigure}
\usepackage[caption=false,font=footnotesize]{subfig}
\usepackage{epsfig}
\usepackage{bm}
\begin{document}
%
\title{On Meta-Learning for Dynamic Ensemble Selection}

\author{\IEEEauthorblockN{Rafael M. O. Cruz and Robert Sabourin}
\IEEEauthorblockA{\'{E}cole de technologie sup\'{e}rieure - Universit\'{e} du Qu\'{e}bec\\
Email: rafaelmenelau@gmail.com, robert.sabourin@livia.etsmtl.ca}
\and
\IEEEauthorblockN{George D. C. Cavalcanti}
\IEEEauthorblockA{Centro de Inform\'{a}tica - Universidade Federal de Pernambuco\\
Email: gdcc@cin.ufpe.br}
}


%

\maketitle

\begin{abstract}

In this paper, we propose a novel dynamic ensemble selection framework using meta-learning. The framework is divided into three steps. In the first step, the pool of classifiers is generated from the training data. The second phase is responsible to extract the meta-features and train the meta-classifier. Five distinct sets of meta-features are proposed, each one corresponding to a different criterion to measure the level of competence of a classifier for the classification of a given query sample. The meta-features are computed using the training data and used to train a meta-classifier that is able to predict whether or not a base classifier from the pool is competent enough to classify an input instance. Three different training scenarios for the training of the meta-classifier are considered: problem-dependent, problem-independent and hybrid. Experimental results show that the problem-dependent scenario provides the best result. In addition, the performance of the problem-dependent scenario is  strongly correlated with the recognition rate of the system. A comparison with state-of-the-art techniques shows that the proposed-dependent approach outperforms current dynamic ensemble selection techniques. 

\end{abstract}


\section{Introduction}

Ensembles of Classifiers (EoC) have been widely studied in the past years as an alternative to increase efficiency and accuracy in many pattern recognition~\cite{kittler,kuncheva}. There are many examples in the literature that show the efficiency of an ensemble of classifiers in various tasks, such as signature verification~\cite{signverification}, handwritten recognition~\cite{Cruz2012,ijcnn} and image labeling~\cite{singh}. Classifiers ensembles involve two basic approaches, namely classifier fusion and dynamic ensemble selection. With classifier fusion approaches, every classifier in the ensemble is used and their outputs are aggregated to give the final prediction. However, such techniques~\cite{kittler,multioraveraging,iwssip,Cruz2012} presents two main problems: they are based on the assumption that the base classifiers commit independent errors, which is difficult to find in real pattern recognition applications. Moreover, not every classifier in the pool of classifiers is an expert for every test pattern. Different patterns are associated with distinct degrees of difficulties. It is therefore reasonable to assume that only a few base classifiers can achieve the correct prediction.

On the other hand, dynamic ensemble selection (DES) techniques work by estimating the level of competence of a classifier for each query sample separately. Then, only the most competent classifiers in relation to the input sample are selected to form the ensemble. Thus, the key point in DES techniques is to define a criterion to measure the level of competence of a base classifier for the classification of the given query sample. In the literature, we can observe several criteria based on estimates of the classifier accuracy in local regions of the feature space surrounding the query sample~\cite{lca,ijcnn2011,knora,Smits_2002,Woloszynski}, extent of consensus~\cite{docs} and decision templates~\cite{mcb,bks,paulo2,outprof}. However, in our previous works~\cite{ijcnn2011}, we demonstrate that using only one criterion to measure the level of competence of a base classifier is very error-prone. 

In this paper, we propose a novel dynamic ensemble selection framework using meta-learning. The framework is divided into three steps: (1) overproduction, where the pool of classifiers is generated, (2) Meta-training where the meta-features are extracted, using the training data, and used as inputs to train a meta-classifier that works as a classifier selector. Five sets of meta-features are proposed in this work. Each set of meta-features correspond to a different criteria used to measure the level of competence of a base classifier such as the confidence of the base classifier for the classification of the input sample, and its performance in predefined regions of the feature space. (3) Generalization phase, in which the meta-features are extracted from each query sample and used as input to the meta-classifier to perform the ensemble selection. Thus, based on the proposed framework we integrate multiple dynamic selection criteria in order to achieve a more robust dynamic selection technique. 

Three different training scenarios for the meta-classifier are investigated: (1) The meta-classifier is trained using data from one classification problem, and is used as the classifier selector~\footnote{In this paper, we use the terms meta-classifier and classifier selector interchangeably} on the same problem; (2) The meta-classifier is trained using one classification problem, and is used as the classifier selector on a different one; (3) A single meta-classifier is trained using the data of all classification problems considered in this work, and is used as the classifier selector for all classification problems. 

Based on these three scenarios, we aim to answer three research questions: (1) Can the use of meta-features lead to a more robust dynamic selection technique?  (2) Is the training of the meta-classifier problem-dependent? (3) Can we improve the performance of the meta-classifier using knowledge from different classification problems? Experiments conducted over eleven classification datasets demonstrate that the proposed technique outperforms current dynamic selection techniques. Furthermore, the accuracy of the DES system is correlated to the performance of the meta-classifier.

This paper is organized as follows: In Section~\ref{sec:relatedWork} we introduce the notion of classifier competence for dynamic selection. The architecture of the proposed system is presented in Section~\ref{sec:proposed}. Experimental results are given in Section~\ref{sec:experiments}. Finally, a conclusion is presented in the last section.

\section{Classifier Competence}
\label{sec:relatedWork}

The level of competence of a classifier defines how much we trust an expert, given a classification task. It is used as a way of selecting, from a pool of classifiers $C$, the one(s) that best fit(s) a given test pattern $\mathbf{x}_{j}$. Thus, in dynamic selection, the level of competence is measured on-the-fly according to some criteria applied for each input instance separately. There are three categories present in the literature~\cite{Alceu2014}: the classifier accuracy over a local region, i.e., in a region close to the test pattern; decision templates, and the extent of consensus.

\subsection{Classifier accuracy over a local region}
\label{sec:localaccuracy}

Classifier accuracy is the most commonly used criterion for dynamic classifier and ensemble selection techniques~\cite{lca,knora,ijcnn2011,clussel,classrank,Smits_2002,dceid}.
Techniques that are based on this paradigm first define a local region around the test instance, called the region of competence. This region is computed using either the K-NN algorithm~\cite{knora,lca,ijcnn2011} or by Clustering techniques~\cite{clussel,anne}. For example, the OLA technique~\cite{lca} selects the classifier that obtains the highest accuracy rate in the region of competence. The Local classifier accuracy (LCA)~\cite{lca} selects the classifier with the highest accuracy in relation to a specific class label and the K-Nearests Oracle (KNORA) technique~\cite{knora} selects all classifiers that achieve a perfect accuracy in the region of competence. The drawback of these techniques is that their performance ends up limited by the algorithm that defines the region of competence~\cite{ijcnn2011}.


\subsection{Decision Templates}
\label{sec:decisionTemplates}

In this class of methods, the goal is also to select patterns that are close to the test sample $\mathbf{x}_{j}$. However, the similarity is computed in the decision space through the concept of decision templates~\cite{decisiontemplates}. This is performed by transforming both the test instance $\mathbf{x}_{j}$ and the validation data into output profiles using the transformation $T$, ($T: \mathbf{x}_{j} \Rightarrow \tilde{\mathbf{x}}_{j}$), where $\mathbf{x}_{j} \in \Re^{D}$ and $\tilde{\mathbf{x}}_{j} \in Z^{M}$~\cite{logid,outprof} ($M$ is the pool size). The output profile of a pattern $\mathbf{x}_{j}$ is denoted by $\tilde{\mathbf{x}}_{j} = \left\lbrace \tilde{\mathbf{x}}_{j,1}, \tilde{\mathbf{x}}_{j,2}, \ldots, \tilde{\mathbf{x}}_{j,M} \right\rbrace $, where each $\tilde{\mathbf{x}}_{j,i}$ is the decision yielded by the classifier $c_{i}$ for $\mathbf{x}_{j}$. Based on the information extracted from the decision space, the K-Nearest Output Profile (KNOP)~\cite{logid} is similar to the KNORA technique, with the difference being that the KNORA works in the feature space while the KNOP works in the decision space. The Multiple Classifier Behaviour (MCB) technique~\cite{mcb} selects the classifiers that achieve a performance higher than a given threshold.  The problem with using such information lies in the fact it neglects the local performance of the base classifiers.

\subsection{Extent of Consensus or confidence}
\label{sec:confidence}

In this class of techniques, the first step is to generate a population of an ensemble of classifiers (EoC), $C^{*} = \{C^{'}_{1},C^{'}_{2}, \ldots, C^{'}_{M'}\}$ ($M^{'}$ is the number of EoC generated) using an optimization algorithm such as a genetic algorithms or greedy search~\cite{greedy,multiga}. Then, for each new query instance $\mathbf{x}_{j}$, the level of competence of each EoC is computed using techniques such as the Ambiguity-guided dynamic selection (ADS), Margin-based dynamic selection (MDS) and Class-strength dynamic selection (CSDS)~\cite{docs,paulo2}. The drawback of these techniques is that they require the pre-computation of EoC, which increases the computational complexity. In addition, the pre-computation of EoC also reduces the level of diversity and the Oracle performance (the Oracle performance is the upper limit performance of an EoC~\cite{kuncheva}) of the pool~\cite{docs}.

\section{Proposed dynamic ensemble selector}
\label{sec:proposed}

\begin{figure*}[!ht]
  
   \begin{center}  	 
       	  \epsfig{file=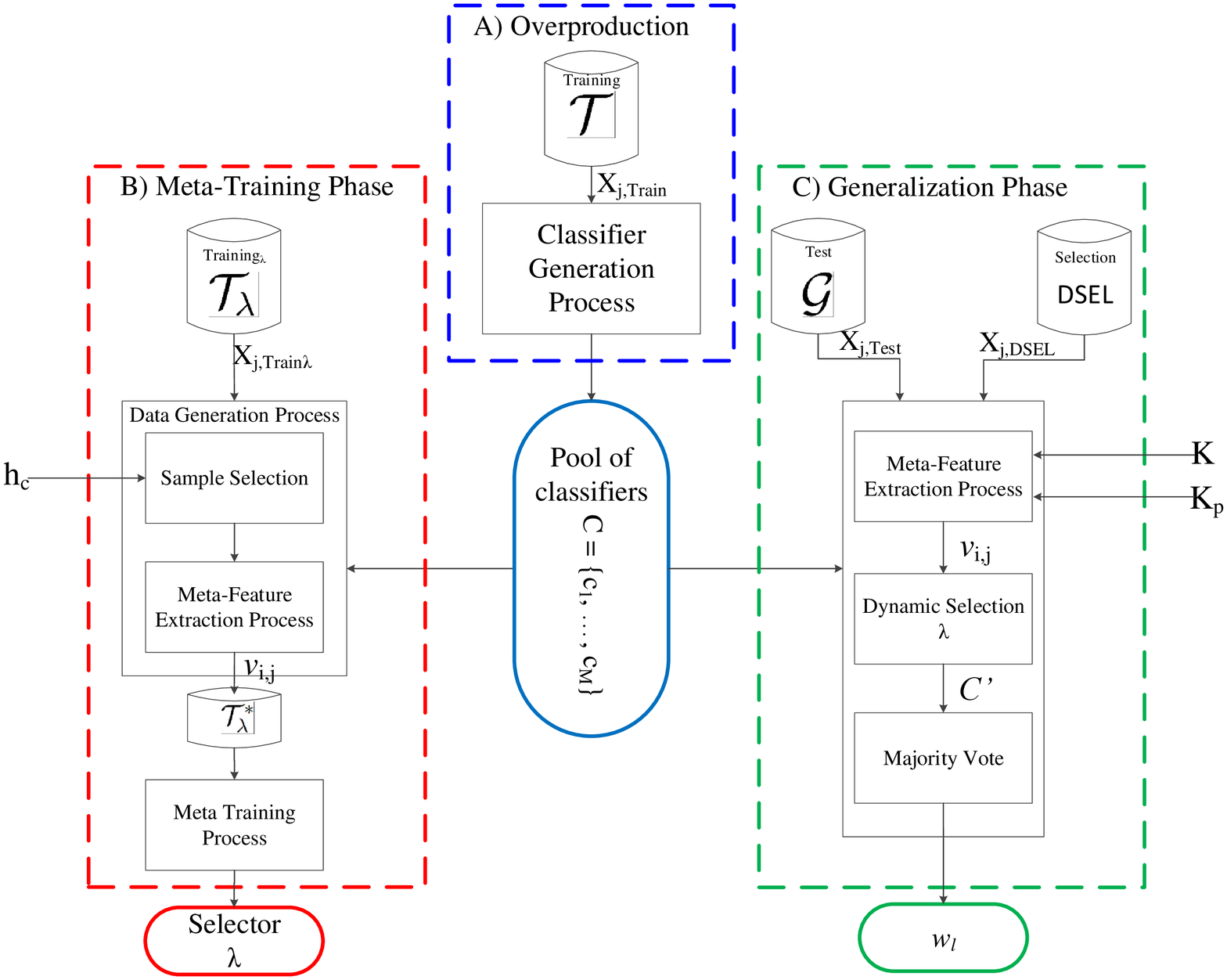, clip=,  width=1.0\textwidth}
   \end{center}
\caption{Overview of the proposed framework. It is divided into three steps 1) Overproduction, where the pool of classifiers $C = \{c_{1}, \ldots, c_{M}\}$ is generated, 2) The training of the selector $\lambda$ (meta-classifier), and 3) The generalization phase where an ensemble $C'$ is dynamically defined based on the meta-information extracted from $\mathbf{x}_{j,test}$ and the pool $C = \{c_{1}, \ldots, c_{M}\}$. The generalization phase returns the label $w_{l}$ of $\mathbf{x}_{j,test}$. $h_{C}$, $K$ and $K_{p}$ are the hyper-parameters required by the proposed system.}
\label{fig:overview}
\end{figure*}

A general overview of the proposed framework is depicted in Figure~\ref{fig:overview}. It is divided into three phases: Overproduction, Meta-training and Generalization.

\subsection{Overproduction} 

In this step, the pool of classifiers $C = \{c_{1}, \ldots, c_{M}\}$, where $M$ is the pool size, is generated using the training dataset $\mathcal{T}$. The Bagging technique~\cite{bagging} is used in this work in order to build a diverse pool of classifiers. 

\subsection{Meta-Training}

In this phase, the meta-features are computed and used to train the meta-classifier $\lambda$. We select five subset of meta-features derived from the three categories presented in Section~\ref{sec:relatedWork}. As shown in Figure~\ref{fig:overview}, the meta-training stage consists of three steps: sample selection, meta-features extraction process and meta-training. A different dataset $\mathcal{T}_{\lambda}$ is used in this phase to prevent overfitting.

\subsubsection{Sample selection}
 
We focus the training of $\lambda$ on cases in which the extent of consensus of the pool is low. Thus, we employ a sample selection mechanism based on a threshold $h_{C}$, called the consensus threshold. For each $\mathbf{x}_{j,train_{\lambda}} \in \mathcal{T}_{\lambda}$, the degree of consensus of the pool, denoted by $H \left ( \mathbf{x}_{j,train_{\lambda}}, C \right )$, is computed. If $H \left ( \mathbf{x}_{j,train_{\lambda}}, C \right )$ falls below the threshold/ $h_{C}$, $\mathbf{x}_{j,train_{\lambda}}$ is passed down to the meta-features extraction process. 

\subsubsection{Meta-feature extraction}
\label{sec:metafeatures}

The first step in extracting the meta-features is to compute the region of competence of $\mathbf{x}_{j,train_{\lambda}}$, denoted by $\theta_{j} = \left \{ \mathbf{x}_{1}, \ldots, \mathbf{x}_{K} \right \}$. The region of competence is defined in the $\mathcal{T_{\lambda}}$ set using the K-Nearest Neighbor algorithm. Then, $\mathbf{x}_{j}$ is transformed into an output profile, $\tilde{\mathbf{x}}_{j}$ by applying the transformation $T$ (Section~\ref{sec:decisionTemplates}). The similarity between $\tilde{\mathbf{x}}_{j}$ and the output profiles of the instances in $\mathcal{T}_{\lambda}$ is obtained through the Manhattan distance. The most similar output profiles are selected to form the set $\phi_{j} = \left \{ \tilde{\mathbf{x}}_{1}, \ldots, \tilde{\mathbf{x}}_{K_{p}} \right \}$, where each output profile $\tilde{\mathbf{x}}_{k}$ is associated with a label $w_{l,k}$. Next, for each base classifier $c_{i} \in C$, five sets of meta-features are calculated:

\begin{itemize}

\item  \emph{\boldsymbol{$f_{1}$} \textbf{- Neighbors' hard classification:}} First, a vector with $K$ elements is created. For each pattern $\mathbf{x}_{k}$, belonging to the region of competence $\theta_{j}$, if $c_{i}$ correctly classifies $\mathbf{x}_{k}$, the $k$-th position of the vector is set to 1, otherwise it is 0. Thus, $K$ meta-features are computed. 

\item \emph{\boldsymbol{$f_{2}$} \textbf{- Posterior Probability:}} First, a vector with $K$ elements is created. Then, for each pattern $\mathbf{x}_{k}$, belonging to the region of competence $\theta_{j}$, the posterior probability of $c_{i}$, $P(w_{l}\mid \mathbf{x}_{k})$ is computed and inserted into the $k$-th position of the vector. Consequently, $K$ meta-features are computed. 

\item \emph{\boldsymbol{$f_{3}$} \textbf{- Overall Local Accuracy:}} The accuracy of $c_{i}$ over the whole region of competence $\theta_{j}$ is computed and encoded as $f_{3}$. 

\item \emph{\boldsymbol{$f_{4}$} \textbf{- Outputs' profile classification:}} First, a vector with $K_{p}$ elements is generated. Then, for each member $\tilde{\mathbf{x}}_{k}$ belonging to the set of output profiles $\phi_{j}$, if the label produced by $c_{i}$ for $\mathbf{x}_{k}$ is equal to the label $w_{l,k}$ of $\tilde{\mathbf{x}}_{k}$, the $k$-th position of the vector is set to 1, otherwise it is set to 0. A total of $K_{p}$ meta-features are extracted using output profiles.  

\item \emph{\boldsymbol{$f_{5}$} \textbf{- Classifier's confidence:}} The perpendicular distance between $\mathbf{x}_{j}$ and the decision boundary of the base classifier $c_{i}$ is calculated and encoded as $f_{5}$. 

\end{itemize}

A vector $v_{i,j} = \left\lbrace f_{1} \cup f_{2} \cup f_{3} \cup f_{4} \cup f_{5} \right\rbrace$ is obtained at the end of the process. If $c_{i}$ correctly classifies $\mathbf{x}_{j}$, the class attribute of $v_{i,j}$, $\alpha_{i,j} = 1$ (i.e., $v_{i,j}$ corresponds to the behavior of a competent classifier), otherwise $\alpha_{i,j} = 0$. $v_{i,j}$ is stored in the meta-features dataset $\mathcal{T}_{\lambda}^{*}$.

\subsubsection{Training}

The last step of the meta-training phase is the training of $\lambda$. The dataset $\mathcal{T}_{\lambda}^{*}$ is divided on the basis of 75\% for training and  25\% for validation. A Multi-Layer Perceptron (MLP) neural network with 10 neurons in the hidden layer is used as the meta-classifier $\lambda$. The training process is stopped if its performance on the validation set decreases or fails to improve for five consecutive epochs.

\subsection{Generalization}

Given an input test sample $\mathbf{x}_{j,test}$ from the generalization dataset $\mathcal{G}$, first, the region of competence $\theta_{j}$ and the set of output profiles $\phi_{j}$, are calculated using the samples from the dynamic selection dataset $D_{SEL}$. For each classifier $c_{i} \in C$, the meta-features are extracted (Section~\ref{sec:metafeatures}), returning the meta-features vector $v_{i,j}$. 

Next, $v_{i,j}$ is passed down as input to the meta-classifier $\lambda$, which decides whether $c_{i}$ is competent enough to classify $\mathbf{x}_{j,test}$. If $c_{i}$ is considered competent, it is inserted into the ensemble $C^{'}$. After each classifier of the pool is evaluated, the majority vote rule~\cite{kuncheva} is applied over the ensemble $C'$, giving the label $w_{l}$ of $\mathbf{x}_{j,test}$. Tie-breaking is handled by choosing the class with the highest a posteriori probability.

\begin{table*}[htbp]
    \centering
    \caption{Mean and standard deviation results of the accuracy for the three scenarios. The best results are in bold. Results that are significantly better ($p < 0.05$) are underlined.}
     \label{table:DESALL} 
     \resizebox{0.80\textwidth}{!}{
      \begin{tabular}{|l| c c c || c c c| }
    \hline
	  
     \textbf{Datasets} &  DES$_{D}$ & DES$_{I}$ & DES$_{ALL}$ &  $\lambda_{D}$  & $\lambda_{I}$ & $\lambda_{ALL}$ \\ 
        \hline
		
        \textbf{Pima} & \textbf{77.74(2.34)} &  72.14(3.69) & 77.18(2.99)  & \textbf{73.20 (3.48)} & 	68.53(1.79)  & 72.57(2.12) \\
        
        \textbf{Liver} & \textbf{\underline{68.83(5.57)}} &  59.22(3.64) &  65.53(3.20) 	 & \textbf{\underline{68.92(2.22)}} & 	52.90(3.66)  & 62.29(3.14)	 \\	
        															 
 		\textbf{Breast} & \textbf{97.41(1.07)} &  96.99(3.64) & 96.96(1.00)	&  \textbf{97.54(1.04)}   & 85.66(6.84)  & 96.97(1.15) \\	
 																		 
 		\textbf{Blood}  & \underline{\textbf{79.14(1.88)}} &   75.39(5.55) &  75.79(2.62)	&  \underline{\textbf{ 82.83(5.57)}} &  	69.32(2.90)  & 74.28(2.87) \\	
 																  
   		\textbf{Banana} & \underline{\textbf{90.16(2.09)}}  & 82.52(13.24) &  85.98(1.73) 	&  \underline{\textbf{91.14(3.09)}}  &  83.58(6.09)  & 80.21(8.97)  \\	
   																		 
		\textbf{Vehicle} & 82.50(2.07) &  	80.25(3.73) &  \underline{\textbf{83.53(1.26)}}	 & 82.38(2.34) &  	73.70(3.85)	 &  \underline{\textbf{88.67(3.15)}} \\	
																  
		\textbf{Lithuanian} & \underline{\textbf{90.26(2.78)}} &  	79.48(13.56) &  87.40(1.87)	&  \underline{\textbf{89.42(3.41)}} &  	82.20(6.31)  & 81.70(3.97)  \\	
													 
		\textbf{Sonar} & 79.72(1.86) &   53.14(6.66) &  \textbf{80.38(4.32)}   & \textbf{76.15(2.43)} &  60.70(7.34)  & 75.42(2.91)  \\	
																
		\textbf{Ionosphere} & \textbf{89.31(0.95)} &  86.69(6.94) &  88.97(2.51)	 & 89.18(2.31) & 67.44(3.42)  & \textbf{89.52(3.72)} \\	
															
		\textbf{Wine} & \underline{\textbf{96.94(3.12})} &   94.39(10.91) & 95.11(6.69)   & \underline{\textbf{93.33(1.56)}} &  	90.86(4.49) & 78.11(6.69)	 \\   
		   														
		\textbf{Haberman} & 76.71(3.52) &  	 72.77(6.34) &  \textbf{77.63(2.55)}	 & \textbf{76.31(2.35)} & 71.88(2.72)  & 76.23(4.91) \\													
 
    \hline
    \end{tabular}
    }
\end{table*}

\section{Experiments}
\label{sec:experiments}

We evaluated the generalization performance of the proposed technique using eleven classification datasets, nine from the UCI machine learning repository, and two, artificially generated using the Matlab PRTOOLS toolbox\footnote[2]{www.prtools.org}. The experiment was conducted using 20 replications. For each replication, the datasets were randomly divided on the basis of 25\% for training ($\mathcal{T}$), 25\% for meta-training $\mathcal{T}_{\lambda}$, 25\% for the dynamic selection dataset ($D_{SEL}$) and 25\% for generalization ($\mathcal{G}$). The divisions were performed maintaining the prior probability of each class. The pool of classifiers was composed of 10 Perceptrons. The value of the hyper-parameters $K$, $K_{p}$ and $h_{c}$ were 7, 5 and 70\% respectively. They were selected empirically based on previous results~\cite{ijcnn2011}. 

We evaluate three different scenarios for the training of the meta-classifier $\lambda$. For the following definitions, let $\mathcal{D} = \left\lbrace \mathcal{D}_{1}, \mathcal{D}_{2}, \ldots, \mathcal{D}_{11} \right\rbrace $ be the eleven classification problems considered in this paper, and $\Lambda = \left\lbrace \lambda_{1}, \lambda_{2}, \ldots, \lambda_{11} \right\rbrace$ a set of meta-classifiers trained using the meta-training dataset, $\mathcal{T}_{\lambda,i}^{*}$ related to a classification problem $\mathcal{D}_{i}$. 

\begin{enumerate}

\item \emph{Scenario I - $\lambda$ dependent($\lambda_{D}$):} The selector $\lambda_{i}$ is trained using the meta-training data $\mathcal{T}_{\lambda,i}^{*}$, and is used as the classifier selector for the same classification problem $\mathcal{D}_{i}$. This scenario is performed in order to answer the first research question of this paper: Can the use of meta-features lead to a more robust dynamic selection technique? 

\item \emph{Scenario II - $\lambda$ independent($\lambda_{I}$):} The selector $\lambda_{i}$ is trained using the meta-training data $\mathcal{T}_{\lambda,i}^{*}$, and is used as the classifier selector for a different classification problem $\mathcal{D}_{j} \mid i \neq j$. The objective of this scenario is to answer the second question posed in this work: Is the training of the meta-classifier application independent?

\item \emph{Scenario III - $\lambda_{ALL}$:} Here, we train a single meta-classifier $\lambda_{ALL}$ using the meta-training data derived from all classification problems $\mathcal{D}_{i} \in \mathcal{D}$, $\mathcal{T}_{\lambda,ALL}^{*} = \left\lbrace \mathcal{T}_{\lambda,1}^{*} \cup \mathcal{T}_{\lambda,2}^{*} \cup \ldots, \cup \mathcal{T}_{\lambda,11}^{*} \right\rbrace$. The objective of this scenario is to answer the third question posed in this paper: Can we improve the performance of the meta-classifier using knowledge from different classification problems?

\end{enumerate}
	
For the rest of this paper, we refer to each scenario as $\lambda_{D}$, $\lambda_{I}$ and $\lambda_{ALL}$. We refer to DES$_{D}$, DES$_{I}$ and DES$_{ALL}$, the DES system created using each training scenario, respectively.
										
\subsection{Results}

\begin{figure}[ht]
   
    \begin{center}  	 
        	  \epsfig{file=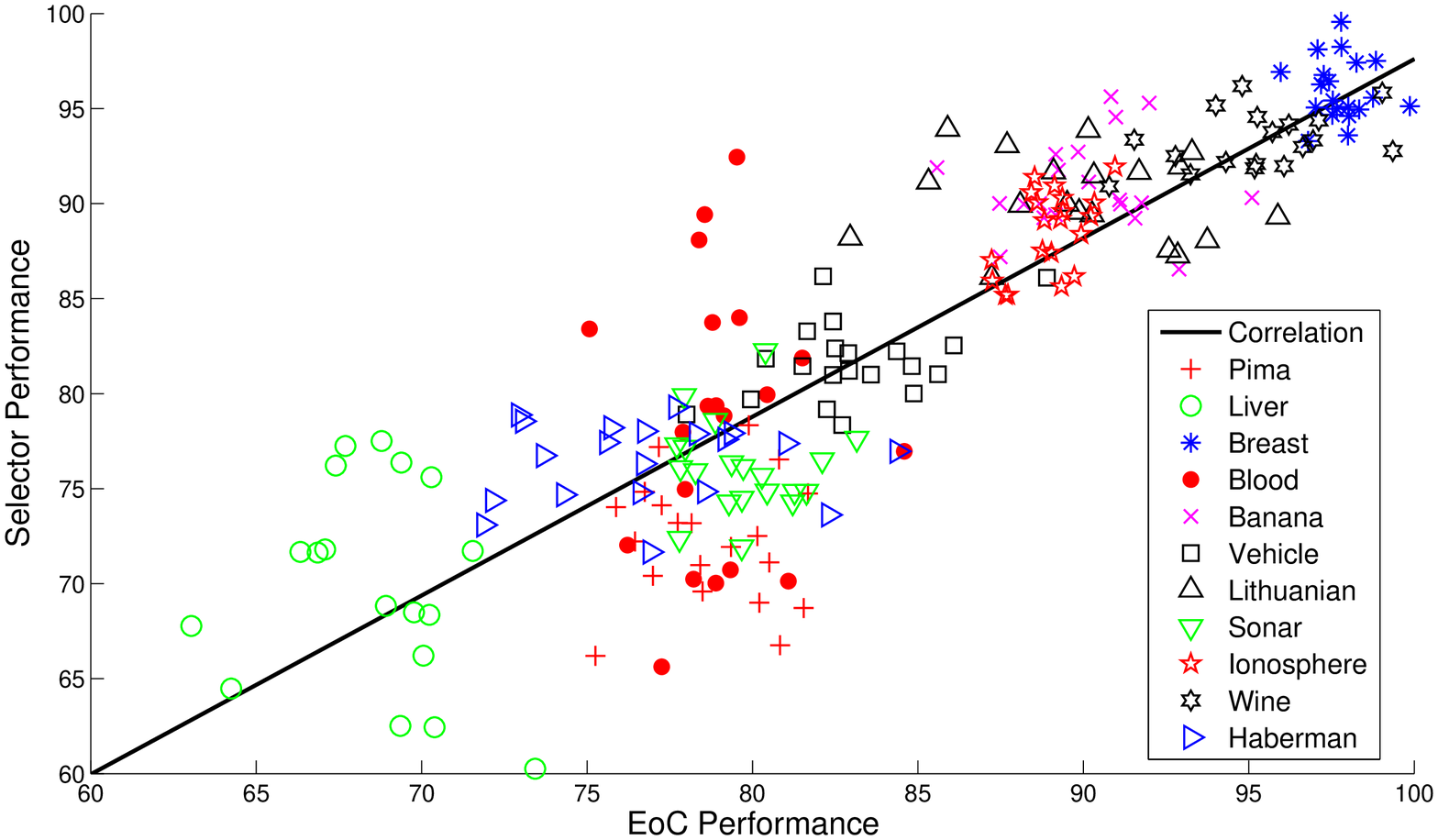, clip=,  width=0.50\textwidth}
    \end{center}
 \caption{Correlation between the performances of the proposed DES$_{D}$ and $\lambda_{D}$. $\rho = 0.88$.}
 \label{fig:correlation}
 \end{figure}

\begin{figure}[ht]
  
\begin{center}  	 
       	  \epsfig{file=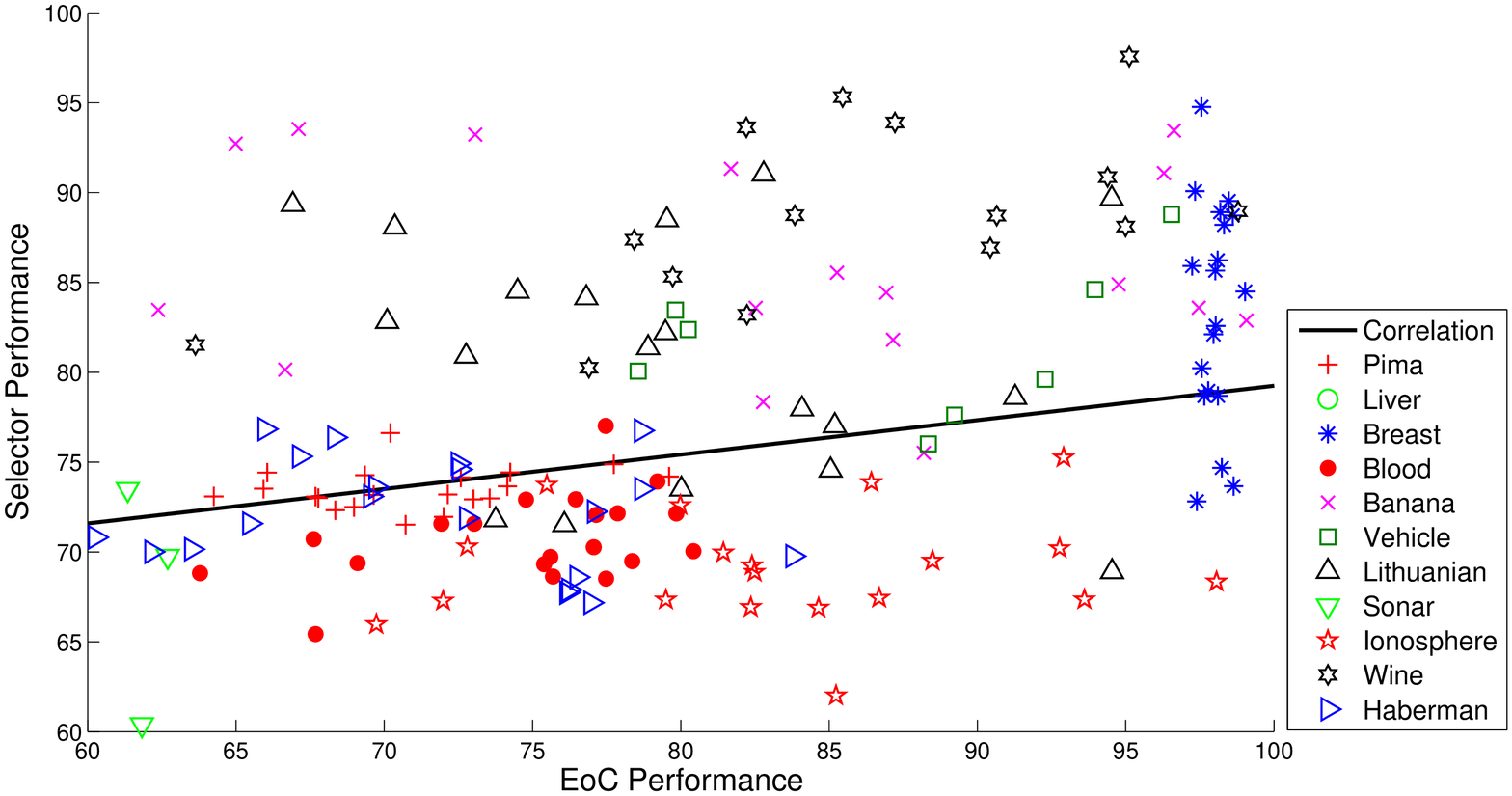, clip=,  width=0.50\textwidth}
   \end{center}
\caption{Correlation between the performances of the proposed DES$_{I}$ and $\lambda_{I}$. $\rho = 0.42$.}
\label{fig:correlationIndependent}
\end{figure}

\begin{figure}[ht]
  
   \begin{center}  	 
       	  \epsfig{file=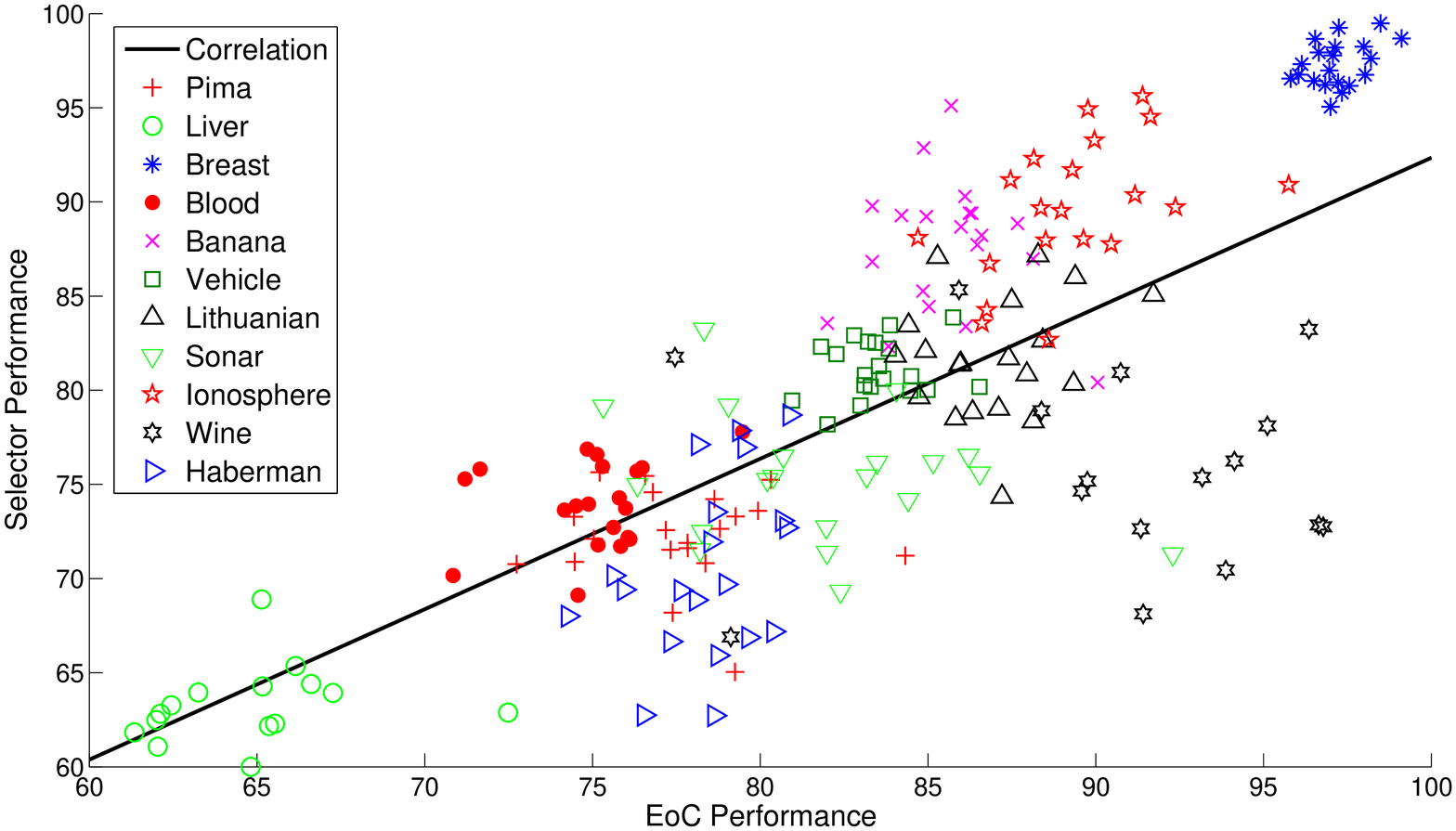, clip=,  width=0.50\textwidth}
   \end{center}
\caption{Correlation between the performances of the proposed DES$_{ALL}$ and $\lambda_{ALL}$. $\rho = 0.76$.}
\label{fig:correlationALL}
\end{figure}

Table~\ref{table:DESALL} shows a comparison of the results achieved related to scenarios I, II and III. Both the DES performance and the meta-classifier performance are presented. We compare each pair of results using the Kruskal-Wallis non-parametric statistical test with a 95\% confidence interval. Results that improved the accuracy significantly are underlined. 

The $\lambda$-dependent scenario (DES$_{D}$) obtained the best results. The only exception is for the Vehicle problem, where the $\lambda_{ALL}$  achieved the best result. Furthermore, when the performance of the meta-classifier is significantly better, the accuracy of the DES system is also significantly better. This finding shows how the performance of the meta-classifier is correlated with the accuracy of its corresponding DES system. 
The independent scenario, $\lambda_{I}$, presented the lowest results for both the DES system (DES$_{I}$) and meta-classifier ($\lambda_{I}$) in all cases. The accuracies of $\lambda_{I}$ and DES$_{I}$ are also significantly worse when compared to the other two scenarios.

We also study the correlation between the accuracy of the DES system and the performance of the meta-classifier for the three scenarios. Figures~\ref{fig:correlation},~\ref{fig:correlationIndependent} and~\ref{fig:correlationALL} show the correlation between the accuracy of the proposed DES system and the performance of the meta-classifier for the $\lambda_{D}$, $\lambda_{I}$ and $\lambda_{ALL}$ scenarios, respectively. We compute the correlation coefficient, $\rho$, using the Pearson's Product-Moment.

Scenario~\textrm{I} achieved the highest correlation coefficient $\rho = 0.88$, while Scenario~\textrm{III} $\lambda_{ALL}$ presented a slightly lower coefficient, $\rho = 0.76$. Thus, the use of knowledge from a different classification problem also reduced the correlation between the meta-classifier and the accuracy of the DES system. The correlation between $\lambda_{I}$ and DES$_{I}$ was $\rho = 0.42$, which is significantly lower than Scenarios~\textrm{I} and~\textrm{III}.

\begin{table*}[ht!]
    \centering
    \caption{Mean and standard deviation results of the accuracy obtained for the proposed DES$_{D}$ and the DES systems in the literature. The best results are in bold. Results that are significantly better ($p < 0.05$) are underlined.}
     \label{table:Results} 
     \resizebox{1.0\textwidth}{!}{
     \begin{tabular}{|c |c |c |c |c |c |c |c|}
    \hline

      \textbf{Database} & \textbf{$DES_{D}$} & \textbf{KNORA-E}~\cite{knora} & \textbf{KNORA-U}~\cite{knora} & \textbf{DES-FA}~\cite{ijcnn2011} & \textbf{LCA}~\cite{lca} & \textbf{OLA}~\cite{lca} & \textbf{KNOP}~\cite{logid} \\
        \hline

        \textbf{Pima} & \underline{\textbf{77.74(2.34)}} & 73.16(1.86) & 74.62(2.18) &  76.04(1.61)  & 72.86(2.98) & 73.14(2.56) & 73.42(2.11) \\

        \textbf{Liver Disorders} & \underline{\textbf{68.92(2.22)}} & 63.86(3.28) & 64.41(3.76) & 65.72(3.81) &  62.24(4.01) &  62.05(3.27) & 65.23(2.29) \\
 
        \textbf{Breast (WDBC)} & \textbf{97.54(1.04)} & 96.93(1.10) & 96.35(1.02) & 97.18(1.13) &  97.15(1.58) & 96.85(1.32) & 95.42(0.89) \\
 
        \textbf{Blood Transfusion} & \underline{\textbf{79.14(1.88)}} & 74.59(2.62) & 75.50(2.36) & 76.42(1.16) &  72.20(2.87) & 72.33(2.36) & 77.54(2.03) \\
      
        \textbf{Banana} & \textbf{90.16(2.09)} & 88.83(1.67) & 89.03(2.87) & \textbf{90.16(3.18)} &  89.28(1.89) & 89.40(2.15) & 85.73(10.65) \\

        \textbf{Vehicle} & \underline{\textbf{82.5(2.07)}} & 81.19(1.54) & 82.08(1.70) & 80.20(4.05) &  80.33(1.84) & 81.50(3.24) & 80.09(1.47)\\

        \textbf{Lithuanian Classes} & 90.26(2.78) & 88.83(2.50) & 87.95(2.64) & \underline{\textbf{92.23(2.46)}} &  88.10(2.20) & 87.95(1.85) & 89.33(2.29) \\
        
        \textbf{Sonar} & \underline{\textbf{79.72(1.86)}} & 74.95(2.79) & 76.69(1.94) & 77.52(1.86) & 76.51(2.06) & 74.52(1.54) & 75.72(2.82) \\
        
        \textbf{Ionosphere} & \underline{\textbf{89.31(0.95)}} & 87.37(3.07) & 86.22(1.67) & 86.33(2.12) & 86.56(1.98) & 86.56(1.98) & 85.71(5.52)\\
        
        \textbf{Wine} & \underline{\textbf{96.94(3.12)}} & 95.00(1.53) & 96.13(1.62) & 95.45(1.77) & 95.85(2.25) & 96.16(3.02) & 95.00(4.14) \\
        
        \textbf{Haberman} & \underline{\textbf{76.71(3.52)}} & 71.23(4.16) & 74.40(2.27) & 74.47(2.41) & 70.16(3.56) & 72.26(4.17) & 75.00(3.40) \\

    \hline
    \end{tabular}
    }
 
\end{table*}

Therefore, experimental results indicate that the training of the meta-classifier is problem-dependent. The behavior of a competent classifier differs according to each classification problem. Furthermore, as the $\lambda_{ALL}$ selector performed worse than the $\lambda_{D}$, we failed to improve the performance of the meta-classifier and DES system by adding knowledge derived from other classification problems. However, the loss in accuracy might be explained by the use of classification problems with completely different distributions and data complexities~\cite{MaciaBOH13}. 

\subsection{Comparison with the state-of-the-art}

In Table~\ref{table:Results}, we compare the recognition rates obtained by the proposed DES$_{D}$ against dynamic selection techniques in the literature (KNORA-Eliminate~\cite{knora}, KNORA-Union~\cite{knora}, DES-FA~\cite{ijcnn2011}, LCA~\cite{lca}, OLA~\cite{lca} and KNOP~\cite{paulo2}). We compare each pair of results using the Kruskal-Wallis non-parametric statistical test with a 95\% confidence interval. The results of the proposed DES$_{D}$ over the Pima, Liver Disorders, Blood Transfusion, Vehicle, Sonar and Ionosphere datasets are statistically superior to the result of the best DES from the literature. For the other datasets, Breast, Banana and Lithuanian, the results are statistically equivalent. 

We can thus answer the first question posed in this paper: Can the use of meta-features lead to a more robust dynamic selection technique? As the result of the proposed DES$_{D}$ is significantly better in eight datasets, the use of meta-learning indeed leads to a more robust dynamic ensemble selection technique. 

\section{Conclusion}

In this paper, we present a novel DES framework using meta-learning. Different properties of the behavior of a base classifier are extracted from the training data and encoded as meta-features. These meta-features are used to train a meta-classifier that can estimate whether a base classifier is competent enough to classify a given input sample. Based on the proposed framework, we perform three experiments considering three different scenarios for the training of the meta-classifier.

Experimental results show that the training of the proposed meta-classifier is problem-dependent as the dependent scenario, $\lambda_{D}$, outperformed both $\lambda_{I}$ and $\lambda_{ALL}$. In addition, the correlation between the performances of $\lambda_{D}$ and the accuracy of the corresponding DES$_{D}$ is also higher than that of the other two scenarios. 

A comparison with the state-of-the-art dynamic ensemble selection techniques shows that the proposed $DES_{D}$ outperforms current techniques. Moreover, the gain in accuracy observed with our system is also statistically significant. Thus, we can conclude that the use of multiple properties of the behavior of a base classifier in the classification environment indeed leads to a more robust DES system.

Future works on this topic will involve:

\begin{enumerate}
\item  The evaluation of a different training scenario using only classification problems with similar data complexity for the training of the meta-classifier.
\item the design of new meta-features in order to improve the performance of the meta-classifier, and consequently, the DES system.
\end{enumerate}

\section*{Acknowledgement}

This work was supported by the Natural Sciences and Engineering Research Council of Canada (NSERC), the \'{E}cole de technologie sup\'{e}rieure (\'{E}TS Montr\'{e}al) and CNPq (Conselho Nacional de Desenvolvimento Cient\'{i}fico e Tecnol\'{o}gico).
 
\bibliographystyle{IEEEbib}
\bibliography{report}

\end{document}